\documentclass[a4paper]{article}
\usepackage{MT-Summit-VII}
\usepackage[dvips]{graphicx}
\usepackage{gb4e-plus}
\usepackage{epsfig}
\usepackage{qobitree}

\title{A Bootstrap Approach to Automatically Generating Lexical
Transfer Rules}

\numauthors{1}

\authori{Davide Turcato \ \ Paul McFetridge \ \ Fred Popowich and
Janine Toole}

\affili{Natural Language Laboratory, School of Computing Science,
Simon Fraser University\\8888 University Drive, Burnaby, British
Columbia, V5A 1S6, Canada\\and\\Gavagai Technology\\P.O. 374, 3495
Cambie Street, Vancouver, British Columbia, V5Z 4R3,
Canada\\\{turk,mcfet,popowich,toole\}@cs.sfu.ca}

\begin{document}

\maketitle

\begin{abstract}
We describe a method for automatically generating Lexical Transfer
Rules (LTRs) from word equivalences using transfer rule
templates. Templates are skeletal LTRs, unspecified for words. New
LTRs are created by instantiating a template with words, provided that
the words belong to the appropriate lexical categories required by the
template. We define two methods for creating an inventory of templates
and using them to generate new LTRs. A simpler method consists of
extracting a finite set of templates from a sample of hand coded LTRs
and directly using them in the generation process. A further method
consists of abstracting over the initial finite set of templates to
define higher level templates, where bilingual equivalences are
defined in terms of correspondences involving phrasal
categories. Phrasal templates are then mapped onto sets of lexical
templates with the aid of grammars. In this way an infinite set of
lexical templates is recursively defined. New LTRs are created by
parsing input words, matching a template at the phrasal level and
using the corresponding lexical categories to instantiate the lexical
template. The definition of an infinite set of templates enables the
automatic creation of LTRs for multi-word, non-compositional word
equivalences of any cardinality.
\end{abstract}

\section{Introduction}

It is well-known that Machine Translation (henceforth MT) systems need
information about the different ways in which words can be translated,
depending on their syntactic and semantic context. Such lexical
transfer rules (henceforth LTRs) are notoriously time-consu\-ming for
humans to construct. Our task is to automatically generate LTRs.

An LTR can be seen as a word equivalence plus an associated transfer
pattern. By \emph{word equivalence} we mean a translation pair simply
stated in terms of words, in a dictionary-like fashion. A transfer
pattern specifies how transfer is to be performed for each of the
morphological variants of the words in the equivalence and for
different syntactic contexts. For instance, given an English-Spanish
word equivalence

\begin{center}
\emph{get lucky} $\leftrightarrow$ \emph{tener suerte}
\end{center}

\noindent the associate transfer pattern would have to account for all
the following equivalences:

\begin{center}
\begin{tabular}{ll}
I \emph{get lucky}			& \emph{Tengo suerte} \\
I \emph{will get lucky}			& \emph{Tendr\'{e} suerte} \\
I \emph{would have got lucky}		& \emph{Habr\'{\i}a tenido suerte} \\
\emph{Getting lucky}			& \emph{Teniendo suerte} \\
I start \emph{getting lucky}		& Empiezo a \emph{tener suerte} \\
I start \emph{getting very lucky}	& Empiezo a \emph{tener mucha} \\
					& \emph{suerte} \\
\end{tabular}
\end{center}

\noindent where the Spanish sentences can be glossed as follows:

\medskip

\begin{tabular}{rll}
a.	& \emph{Tengo}		& \emph{suerte}							\\
	& {I have}		& {good luck}							\\
\end{tabular}

\begin{tabular}{rll}
b.	& \emph{Tendr\'{e}}	& \emph{suerte}							\\
	& {I will have}		& {good luck}							\\
\end{tabular}

\begin{tabular}{rlll}
c.	& \emph{Habr\'{\i}a}	& \emph{tenido}	& \emph{suerte}					\\
	& {I would have}	& had		& {good luck}					\\
\end{tabular}

\begin{tabular}{rll}
d.	& \emph{Teniendo}	& \emph{suerte}							\\
	& Having		& {good luck}							\\
\end{tabular}

\begin{tabular}{rllll}
e.	& Empiezo		& a		& \emph{tener}	& \emph{suerte}			\\
	& {I start}		& to		& have		& {good luck}			\\
\end{tabular}

\begin{tabular}{rlllll}
f.	& Empiezo		& a		& \emph{tener}	& \emph{mucha}	& \emph{suerte} \\
	& {I start}		& to		& have		& much		& {good luck}	\\
\end{tabular}

\medskip

The last example in the list of translation pairs shows that a
transfer pattern also has to account for modifiers. In the example,
\emph{very} is translated by the adjective \emph{mucha}, whereas in a
sentence like \emph{I start getting very lazy} it would be translated
by the adverb \emph{muy} ({\em Empiezo a volverme muy perezoso}).

Thus, a bilingual lexicon of such transfer rules is a different object
from a collection of word equivalences, which is the definition of a
bilingual lexicon most often found in the literature about automatic
creation of bilingual lexicons (\cite{Fung:AMTA98},
\cite{Melamed:AMTA98}, to cite recent examples). As a matter of
fact, those techniques and the one described here are disjoint and
complementary, as the output of those tools can be used as input to
LTR development. Given a collection of word equivalences, we focus on
how transfer patterns can be associated with them to create complete
LTRs.

This task has rarely been tackled before. Several techniques have been
proposed for the automatic acquisition of word equivalences (see
references above), but very few for the automatic acquisition of full
LTRs (e.g. \cite{Copestake:UCCL92}), despite the high cost of their
manual development. Bilingual coding is often a bottleneck in MT
system development. Unlike other linguistic resources, like grammars,
lexicons have an open-ended linear growth and their quality can be
directly related to their size. For this reason, there is often a
mismatch between the development time frames of bilingual lexical
resources and other modules.

\section{Basic ideas}

\subsection{Template based generation}

We use a bootstrap approach to transfer rule creation. An initial hand
coded bilingual lexicon is used as a basis for defining a set of
transfer rule \emph{templates}, i.e. skeletal rules unspecified for
words. Subsequently, appropriate transfer rule templates are
associated to new word equivalences, on the basis of the
morphosyntactic features of those words, to construct complete
LTRs. The approach described here shares this underlying
template-based bootstrap philosophy with the approach described by
\cite{Turcato:COLING98}, but differs from it in three key respects:
the resources it uses, the way templates are created and the way LTRs
are created from word equivalences and templates. LTR templates are
also akin to \emph{tlinks}, as described in \cite{Copestake:UCCL92}
and \cite{Copestake:AAAI93}.These works describe how to use tlinks
for the semi-automatic generation of single-word
equivalences. However, they do not deal with the creation of an
inventory of tlink types or the generation from multi-word
equivalences.

For the sake of exposition, we use here a simplified version of LTRs
and templates, showing only words (for LTRs), syntactic categories and
indices, the latter represented by tuples of subscript lowercase
letters. A schematic LTR and template are shown in (\ref{schemltr})
and (\ref{schemt}), respectively. For a description of the MT system
and the full LTR formalism see \cite{Popowich:TMI97} and
\cite{Turcato:RANLP97}.

\begin{exe}

\ex \label{schemltr} \texttt{\emph{W}$^{s1}$:Cat$^{s1}_{Indices^{s1}}$
\& \ldots \& \emph{W}$^{sm}$:Cat$^{sm}_{Indices^{sm}}$ \\
$\leftrightarrow$ \\ \emph{W}$^{t1}$:Cat$^{t1}_{Indices^{t1}}$ \&
\ldots \& \emph{W}$^{tn}$:Cat$^{tn}_{Indices^{tn}}$}

\ex \label{schemt} \texttt{Cat$^{s1}_{Indices^{s1}}$ \& \ldots \&
Cat$^{sm}_{Indices^{sm}}$ \\ $\leftrightarrow$ \\
Cat$^{t1}_{Indices^{t1}}$ \& \ldots \& Cat$^{tn}_{Indices^{tn}}$}

\end{exe}

Given a word equivalence expressing a translation pair, our goal is to
create a transfer rule directly usable by an MT system. In other
words, the goal is to associate a transfer pattern, as informally
described above, to a word equivalence. We describe two approaches,
the latter of which is an extension of the former.

\subsection{The enumerative approach}

The goal of creating templates and using them in generating LTRs can
be accomplished through the following steps:

\begin{enumerate}

\item Create transfer rule templates:

\begin{enumerate}

\item Define a set of LTR templates. Each template represents a
transfer pattern. The template definition task is carried out by
extracting LTR templates from an initial hand-coded bilingual
lexicon. This can be easily done by simply removing words from LTRs,
normalizing variables by renaming them in some canonical way (so as to
avoid two instances of the same templates to only differ by variable
names), then ranking templates by frequency, if the application of
some cutoff is in order. Table \ref{coverage} shows the incremental
coverage of the set of templates we extracted from our initial
hand-coded English-Spanish bilingual lexicon. We refer to the template
creation process described here as the \emph{enumerative} approach to
building templates.

\begin{table}[tb]
\begin{center}
\begin{tabular}{rrr}
\hline
Templates	& LTRs 		& Coverage	\\
\hline
1		& 5683		& 33.9 \%	\\
2		& 8726		& 52.1 \%	\\
3		& 10710		& 63.9 \%	\\
4		& 12336		& 73.6 \%	\\
5		& 13609		& 81.2 \%	\\
50		& 15473		& 92.3 \%	\\
500		& 16338		& 97.5 \%	\\
922		& 16760		& 100.0 \%	\\
\hline
\end{tabular}
\end{center}
\caption{\label{coverage}Incremental template coverage}
\end{table}

\item Associate a set of constraints to each template. Typically,
these are morphosyntactic constraints on the words to be matched
against the template. Basically, such constraints ensure that an input
word belongs to the same lexical category of the template item it has
to match. The same goal could also be achieved by directly unifying a
lexical category associated to a word with the corresponding template
item, instead of having separate constraints.

\end{enumerate}

\item Create transfer rules:

\begin{enumerate}

\item Given a word equivalence, create an LTR if the lexical
descriptions of the words in the translation pair satisfy all the
constraints associated with a template (or unify with their
corresponding items in the template). In that case, an LTR is created
by simply instantiating the successful template with the words in the
word equivalence.

\end{enumerate}

\end{enumerate}

An enumerative approach to template creation \linebreak guarantees an
adequate coverage for creating most of the LTRs needed in an MT
system, as discussed in \cite{Turcato:COLING98}. A simple LTR
automatic generation procedure can be implemented by selecting the
most significant templates in the database. This can be done, as
hinted above, by counting the occurrences of each template in the LTR
corpus and choosing those that rank best. The top ranking templates
are then used to directly map input word equivalences onto LTRs. This
approach was implemented and used, with good results (e.g. in an early
test run on a 1544 entry word-list downloaded from the World Wide Web,
LTRs were created for 79\% of the input word equivalences. Further
results are discussed in section \ref{performance-sec}.

\subsection{The generative approach}

The idea of adding recursion to the template definition procedure,
thus replacing a finite set of templates with an infinite one, was
brought about by work on phrasal verbs. Phrasal verbs exhibit a larger
variability than other collocations. One of the problems is that their
translations are often paraphrases, because a target language might
lack a direct equivalent to a source phrasal verb. Table \ref{sit}
illustrates this point (e.g. \emph{sit through something}
$\leftrightarrow$ \emph{permanecer hasta la fin de algo}).

\begin{table*}[tbp]

Phrasal verb equivalences:

\begin{center}

\begin{tabular}{llll}
\emph{sit back}			& $\leftrightarrow$ \emph{ponerse c\'{o}modo}			& \texttt{iv adv}	& $\leftrightarrow$ \texttt{rv adj}		\\
\emph{sit down}			& $\leftrightarrow$ \emph{sentarse}				& \texttt{iv adv}	& $\leftrightarrow$ \texttt{rv}			\\
\emph{sit for} sth		& $\leftrightarrow$ \emph{posar para} algo			& \texttt{iv p}		& $\leftrightarrow$ \texttt{iv p}		\\
\emph{sit in for} sb		& $\leftrightarrow$ \emph{sustituir} algn			& \texttt{iv adv p}	& $\leftrightarrow$ \texttt{tv}			\\
\emph{sit in on} sth		& $\leftrightarrow$ \emph{participar como observador en} algo	& \texttt{iv adv p}	& $\leftrightarrow$ \texttt{iv p n p}		\\
\emph{sit through} sth		& $\leftrightarrow$ \emph{permanecer hasta la fin de} algo	& \texttt{iv p}		& $\leftrightarrow$ \texttt{iv p det n p}	\\
\emph{sit tight}		& $\leftrightarrow$ \emph{esperar}				& \texttt{iv adj}	& $\leftrightarrow$ \texttt{iv}			\\
\emph{sit up with} sb		& $\leftrightarrow$ \emph{velar} algn				& \texttt{iv adv p}	& $\leftrightarrow$ \texttt{tv}			\\
\emph{sit up}			& $\leftrightarrow$ \emph{incorporarse}				& \texttt{iv adv}	& $\leftrightarrow$ \texttt{rv}			\\
\emph{sit} sb \emph{down}	& $\leftrightarrow$ \emph{sentar} algn				& \texttt{tv adv}	& $\leftrightarrow$ \texttt{tv} 		\\
\emph{sit} sb \emph{up}		& $\leftrightarrow$ \emph{incorporar} algn			& \texttt{tv adv}	& $\leftrightarrow$ \texttt{tv}			\\
\emph{sit} sth \emph{out}	& $\leftrightarrow$ \emph{no participar en} algo		& \texttt{tv adv}	& $\leftrightarrow$ \texttt{neg iv p}		\\
\end{tabular}

\end{center}

Glosses for Spanish:

\begin{tabular}{rrll}
\hspace{3.7cm}	& a.	& \emph{ponerse} 	& \emph{c\'{o}modo}									\\
		&	& put oneself		& comfortable										\\
\end{tabular}

\begin{tabular}{rrl}
\hspace{3.7cm}	& b.	& \emph{sentarse}												\\
		&	& sit oneself													\\
\end{tabular}

\begin{tabular}{rrlll}
\hspace{3.7cm}	& c.	& \emph{posar}		& \emph{para}		& algo								\\
		&	& pose			& for			& sth								\\
\end{tabular}

\begin{tabular}{rrll}
\hspace{3.7cm}	& d.	& \emph{sustituir}	& algn											\\
		&	& replace		& sb											\\
\end{tabular}

\begin{tabular}{rrlllll}
\hspace{3.7cm}	& e.	& \emph{participar}	& \emph{como}		& \emph{observador}	& \emph{en}	& algo			\\
		&	& take part		& as			& observer		& in		& sth			\\
\end{tabular}

\begin{tabular}{rrllllll}
\hspace{3.7cm}	& f.	& \emph{permanecer}	& \emph{hasta}		& \emph{la}		& \emph{fin}	& \emph{de}	& algo	\\
		&	& remain		& until			& the			& end		& of		& sth	\\
\end{tabular}

\begin{tabular}{rrl}
\hspace{3.7cm}	& g.	& \emph{esperar}												\\
		&	& wait														\\
\end{tabular}

\begin{tabular}{rrll}
\hspace{3.7cm}	& h.	& \emph{velar}		& algn											\\
		&	& watch over		& sb											\\
\end{tabular}

\begin{tabular}{rrl}
\hspace{3.7cm}	& i.	& \emph{incorporarse}												\\
		&	& raise oneself													\\
\end{tabular}

\begin{tabular}{rrll}
\hspace{3.7cm}	& j.	& \emph{sentar}		& algn											\\
		&	& sit			& sb											\\
\end{tabular}

\begin{tabular}{rrll}
\hspace{3.7cm}	& k.	& \emph{incorporar}	& algn											\\
		&	& raise			& sb											\\
\end{tabular}

\begin{tabular}{rrllll}
\hspace{3.7cm}	& l.	& \emph{no}		& \emph{participar}	& \emph{en}		& algo					\\
		&	& do not		& take part		& in			& sth					\\
\end{tabular}

\begin{tabular}{lll}
where: &&\\
En. sth	& = Sp. algo	& = something	\\
En. sb	& = Sp. algn	& = somebody	\\
\end{tabular}
\caption{Some phrasal verb equivalences for the verb \emph{sit}, and associated glosses for Spanish.}\label{sit}
\end{table*}

A finite set of templates could still be used for phrasal verbs, but
this would require a much larger initial LTR corpus than is necessary
for other collocations, in order to preserve a high automatic
generation rate. An alternative solution is based on introducing a
further level of abstraction, by defining higher level, underspecified
templates which state bilingual equivalences in terms of phrasal
categories instead of lexical categories. Then, a simple grammar is
used to map such phrasal categories onto sets of lexical categories in
order to derive completely specified templates.

Despite the template variability in terms of sequences of syntactic
categories, a much higher regularity can be found by defining
templates in terms of constituency. For instance, all the lexical
equivalences listed for \emph{sit} in Table \ref{sit} can be reduced
to two basic patterns in terms of phrasal categories:\footnote{We
adopt the convention of using lowercase labels for lexical categories
and uppercase labels for phrasal categories.}

\begin{exe}

\ex \texttt{VP} $\leftrightarrow$ \texttt{VP}

\ex \texttt{VP/NP} $\leftrightarrow$ \texttt{VP/NP}

\end{exe}

\noindent where \texttt{VP/NP} represents a verbal phrase with a noun
phrase gap. A further generalization is that a \texttt{VP} on either side
of a template tends to be equivalent to a phrase of the same type,
with the same number and type of gaps. We note incidentally that this
further generalization does not hold for all categories, in terms of
category identity. For example, an English adjective often corresponds
to a Spanish prepositional phrase (e.g. \emph{fashionable}
$\leftrightarrow$ \emph{de moda}, \emph{stainless} $\leftrightarrow$
\emph{sin tacha}).

The abstraction process consists of partitioning lexical templates
into classes such that each class is identified by a phrasal template,
in which a group of lexical categories is replaced by a phrasal
category. All the lexical templates in a class can be obtained by
replacing the phrasal category with one of its lexical
projections. Such replacement can be carried out on a purely
monolingual basis, by using a simple grammar to define
constituency. The key requirement on the abstraction process is that
the resulting abstract template be invariant with respect to lexical
replacement, i.e. the replacement do not involve any other element in
the abstract template, beside the replaced phrasal category. This
restriction amounts to requiring that a phrasal category be
self-contained in terms of variable sharing, i.e. the lexical
categories it dominates introduce no new variables to be shared with
items external to the phrase itself; or, if such sharing happens, it
must be entirely predictable and unambiguous, i.e. the variable sharing
lexical category must be marked in such a way that it can be uniquely
identified.

\begin{figure*}[tb]
\begin{center}
\epsfig{file=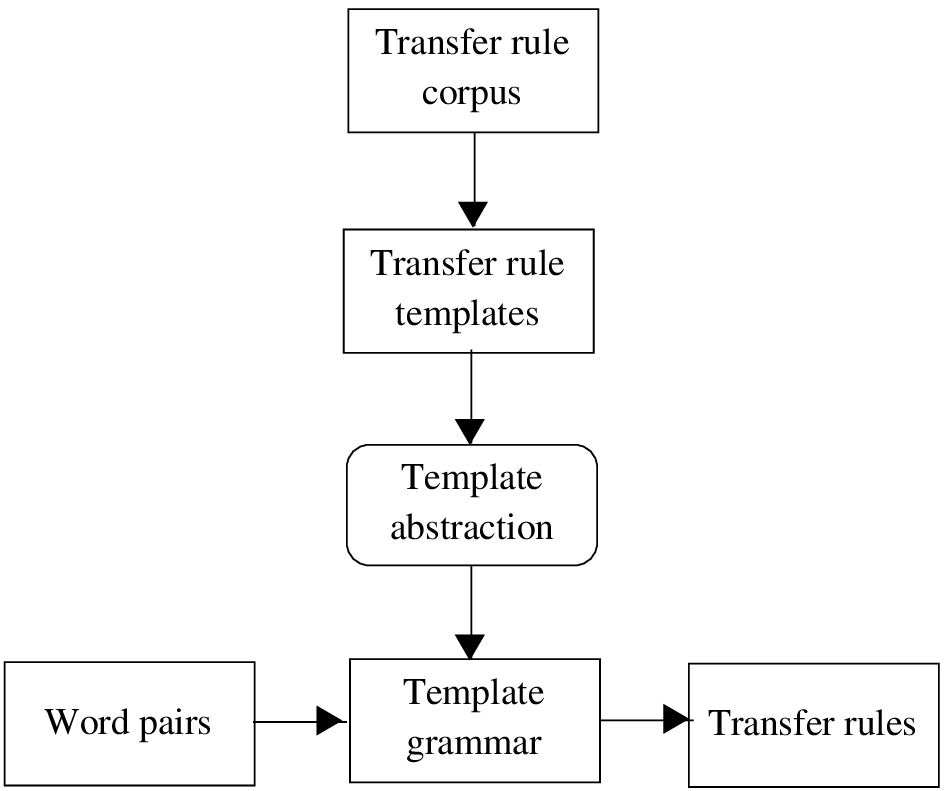}
\caption{System architecture.}\label{architecture}
\end{center}
\end{figure*}

Once a phrasal template has been defined, and a grammar is available
for mapping phrasal categories onto lexical categories, new,
previously unseen lexical templates can be derived that were not
present in the initial class which the phrasal template was abstracted
over. Depending on the recursivity of the grammars in use, an infinite
set of lexical templates can be defined via a finite set of phrasal
template. Therefore, we refer to this process as to a
\emph{generative} approach to building templates. For example, the
templates underlying the bilingual entries in (\ref{buddha}) allow one
to infer the template underlying the bilingual entry for
(\ref{halloween}).

\begin{exe}

\ex \label{buddha} \emph{Buddha} $\leftrightarrow$ \emph{buda}

\emph{Wonderland} $\leftrightarrow$ \emph{pa\'{\i}s de las maravillas}

\ex \label{halloween} \emph{Halloween} $\leftrightarrow$
\emph{v\'{\i}spera del D\'{\i}a de los Santos}

\end{exe}

Namely, from the lexical templates in (\ref{buddhaltr}), the phra\-sal
template in (\ref{buddhat}) can be inferred. In turn, the new lexical
template in (\ref{halloweenltr}) can be derived from (\ref{buddhat}),
provided that the relevant grammar licenses the projection of a
phrasal category \texttt{NBAR} onto an appropriate sequence of lexical
categories.

\begin{exe}

\ex \label{buddhaltr} 

\texttt{n$_{a}$ $\leftrightarrow$ n$_{a}$}

\texttt{n$_{a}$ $\leftrightarrow$ n$_{a}$ \& p$_{a,b}$ \& d$_{b}$ \& n$_{b}$}

\ex \label{buddhat}

\texttt{n$_{a}$ $\leftrightarrow$ NBAR$_{a}$}
\ex \label{halloweenltr}

\texttt{n$_{a}$ $\leftrightarrow$ n$_{a}$ \& p$_{a,b}$ \& d$_{b}$ \& n$_{b}$ \&
p$_{b,c}$ \& d$_{c}$ \& n$_{c}$}

\end{exe}

The overall architecture of the generative approach is shown in Figure
\ref{architecture}. The idea of the whole process is to exploit
monolingual regularities to account for a non-compositional bilingual
equivalence. In a non-compositional equivalence, direct
correspondences between lexical items on either side cannot be
established. The two sides are only equivalent as wholes. However, a
non-compositional equivalence can be accounted for by decomposing it
into a phrasal bilingual equivalence and monolingual mappings from
phrasal to lexical categories.

Implementing a template grammar allows one to obtain an adequate
template coverage while requiring only a small initial LTR corpus. In
our system the template abstraction task was performed
manually. Although the implementation of an automatic template
abstraction procedure could be foreseen, the high reliability required
of templates demands that a strict human control still be placed at
some point of the template abstraction and grammar definition
phases. It is also worth pointing that, by our experience, the grammar
development task is not very labour intensive. Such grammars have to
perform a very limited task, and deal with a restricted and controlled
input. Basically, they only have to account for the constituent
structure of very simple and syntactically ordinary phrases. In our
case, the grammar development time was usually measured in hours.

\section{Implementation}

\begin{figure*}[tb]
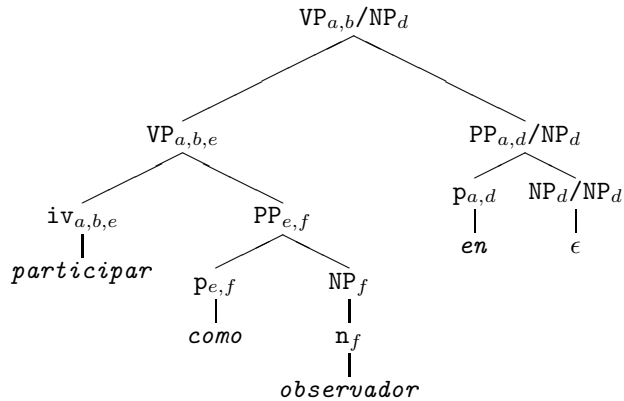

\texttt{
\leaf{\emph{participar}}
\branch{1}{iv$_{a,b,e}$}
\leaf{\emph{como}}
\branch{1}{p$_{e,f}$}
\leaf{\emph{observador}}
\branch{1}{n$_{f}$}
\branch{1}{NP$_{f}$}
\branch{2}{PP$_{e,f}$}
\branch{2}{VP$_{a,b,e}$}
\leaf{\emph{en}}
\branch{1}{p$_{a,d}$}
\leaf{$\epsilon$}
\branch{1}{NP$_{d}$/NP$_{d}$}
\branch{2}{PP$_{a,d}$/NP$_{d}$}
\branch{2}{VP$_{a,b}$/NP$_{d}$}
\tree
}
\caption{Template right hand side generation tree.}\label{tree}
\end{figure*}

In this section the generation of an LTR from a phrasal template is
described. We illustrate the procedure with the aid of a worked
example. We show how the LTR in (\ref{ltr}) is generated from the word
equivalence in (\ref{we}).

\begin{exe}

\ex \label{we} \emph{sit in on} sth $\leftrightarrow$ \\
\emph{participar como observador en} algo

\end{exe}

\begin{exe}

\ex \label{ltr}
\texttt{\emph{sit}:iv$_{a,b,c}$ \& \emph{in}:adv$_{c}$ \& \emph{on}:p$_{a,d}$
$\leftrightarrow$\\
\emph{participar}:iv$_{a,b,e}$ \& \emph{como}:p$_{e,f}$ \& \\
\emph{observador}:n$_{f}$ \& \emph{en}:p$_{a,d}$
}

\end{exe}

Pragmatic reasons induced us to take a hybrid approach to the template
grammar construction task, combining the enumerative and generative
approaches to build templates. It turned out that templates show a
larger variability on the Spanish side than on the English side. Table
\ref{sit} is a good example of this point. This fact might be related
to a larger use of concise and productive phrasal verb patterns in
English, or perhaps to the fact that our word equivalence coding was
driven by English collocations, thus involving the presence of
paraphrases on the Spanish side only. In any case it appeared that the
enumerative approach was sufficient to adequately cover the English
side without resulting in a proliferation of templates. Hence, we
constructed a set of templates in which only the Spanish side
contained a phrasal category, while the English side was fully
specified. For example:

\begin{exe}
\ex \label{phrasal-template} 
\texttt{iv$_{a,b,c}$ 
\& adv$_{c}$ 
\& p$_{a,d}$
$\leftrightarrow$ 
VP$_{a,b}$/NP$_{d}$
}
\end{exe}

For each such partially specified template, the Spanish side has to be
generated.

The selection of a phrasal template candidate is performed by doing a
lexical lookup for the source words and matching their morphosyntactic
representations with the corresponding left hand side template
items. This can be done directly by unification or by associating a
set of constraints to the phrasal template. In our implementation,
constraints are expressed by Prolog goals taking morphosyntactic
descriptions as arguments. For example, the phrasal template in
(\ref{phrasal-template}) is selected as a candidate for our input
translation pair in (\ref{we}), as the lexical descriptions of
\emph{sit}, \emph{in} and \emph{on} match the categories \texttt{iv},
\texttt{adv} and \texttt{p}, respectively. A sequence of words can
match several phrasal templates, e.g. in our specific example the
following candidate is also selected, since \emph{in} can also be a
noun (e.g. \emph{the ins and outs of a problem}):

\begin{exe}
\ex 
\texttt{tv$_{a,b,c}$ 
\& n$_{c}$ 
\& p$_{c,d}$
$\leftrightarrow$ 
VP$_{a,b}$/NP$_{d}$
}
\end{exe}

Given a candidate phrasal template, the core of the generation
procedure is a call to a target language grammar, Spanish in this
case, which parses the target input words using the given phrasal
category (with its associated indices) as its initial symbol.

A phrasal template may disjunctively specify several phrasal
categories as initial symbols (or, equivalently, different phrasal
templates may share the same English side, while specifying different
initial symbols). For instance, English adjectives may be equivalent
to either a Spanish adjectival phrase or prepositional phrase, as
already mentioned.

Figure \ref{tree} shows the result of parsing the Spanish input words
of our example, using the initial symbol in the phrasal template in
(\ref{phrasal-template}). Each node shows the assigned syntactic
category, along with the indices, either specified in the initial
symbol or instantiated during parsing.

The pre-terminal categories in the parse tree, along with the input
words, are used to build the following LTR right hand side for the
final LTR:

\begin{exe}
\ex \label{rhs} 
\texttt{\emph{participar}:iv$_{a,b,e}$ \& \emph{como}:p$_{e,f}$ \& \\
\emph{observador}:n$_{f}$ \& \emph{en}:p$_{a,d}$
}
\end{exe}

Finally, by instantiating the left hand side of the template with the
input English words and replacing the right hand side phrasal category
in the phrasal template (\ref{phrasal-template}) with the right hand
side in (\ref{rhs}), the LTR in (\ref{ltr}) is obtained.

Again, several output LTRs can be generated for the same input
words. For example, the following LTRs are also generated for our
example:

\begin{exe}
\ex \label{altltr1} \texttt{\emph{sit}:iv$_{a,b,c}$ 
\& \emph{in}:adv$_{c}$ 
\& \emph{on}:p$_{a,d}$ 
$\leftrightarrow$\\
\emph{participar}:iv$_{a,b}$ \& \emph{como}:p$_{a,e}$ \& \\
\emph{observador}:n$_{e}$ \& \emph{en}:p$_{a,d}$ 
}
\ex \label{altltr2} \texttt{\emph{sit}:iv$_{a,b,c}$ 
\& \emph{in}:adv$_{c}$ 
\& \emph{on}:p$_{a,d}$ 
$\leftrightarrow$\\
\emph{participar}:iv$_{a,b,e}$ \& \emph{como}:p$_{e,f}$ \& \\
\emph{observador}:n$_{f}$ \& \emph{en}:p$_{f,d}$
}
\end{exe}

In (\ref{altltr1}) the indices show that \emph{como} is analyzed as a
modifier of \emph{participar}, instead of a complement. In
(\ref{altltr2}) the preposition \emph{en} modifies \emph{observador},
instead of \emph{participar}. In our case, since we are interested in
translating only from English to Spanish, we would keep only one of
the candidate LTRs shown above, as they all share the same English
side. However, given that the LTRs at hand are bidirectional, if one
were interested in translating from Spanish to English, one might want
to keep all the entries, in order to cover different syntactic
analyses.

\begin{table*}[tb]
\begin{center}
\begin{tabular}{lrrrrrc}
File			& In	& Out	& Val	& InOut	& InVal	& \%		\\ \hline
\textbf{Enumerative approach:} \\
ADJ			& 542	& 546	& 469	& 468	& 468	& 86.3 \%	\\
Phrasal verbs - batch 1	& 2340	& 2395	& 1416	& 1647	& 1414	& 60.4 \%	\\ \hline
\textbf{Generative approach:} \\
Phrasal verbs - batch 2	& 549	& 1152	& 486	& 512	& 469	& 85.4 \%	\\
Phrasal verbs - batch 3	& 478	& 782	& 404	& 418	& 393	& 82.2 \%	\\
V + (ADJ or N)		& 345	& 617	& 300	& 302	& 292	& 84.6 \%	\\
ADJ + N			& 1144	& 1331	& 914	& 914	& 903	& 78.9 \%	\\
IV + ADJ		& 199	& 346	& 187	& 187	& 187	& 94.0 \%	\\
\end{tabular}
\end{center}
\caption{Overview of system performance.}\label{performance}
\end{table*}

The last step is the validation of LTRs by lexicographers, which
exclusively consists of removing unwanted entries. Lexicographers use
their linguistic intuition and knowledge of the syntactic
representations used in LTRs, in order to make a choice among several
competing LTRs for a given translation pair, or to to check the
correctness of the analysis underlying an LTR. Information that they
are typically expected to check is the way coindexing is performed
(e.g. for prepositions, in order to check that they be attached to the
correct item) and the syntactic categories assigned to words, most
crucially when unknown words are involved. In our implementation
lexicographers are helped in the task by messages that the system
associates to candidate LTRs, in order to signal, for instance, the
presence of lexically unknown words or lexical ambiguities which are
potential sources of errors (for instance, verbs which are both
transitive and intransitive). The files output by the system tend to
be self-contained in terms of information needed by lexicographers to
perform their task, and usually lexicographers do not need access to
any extra linguistic resources. We finally note that the validation
step becomes particularly crucial if the input translation pairs are
automatically acquired from resources like bilingual corpora, in order
to filter out LTRs created from noisy bilingual equivalences.

\section{Performance}\label{performance-sec}

Results of this approach are shown in Table \ref{performance}. Columns
are to be interpreted as follows:

\begin{description}

\item[File]: The type of words or collocations in the processed
file.

\item[In]: The number of input word equivalences.

\item[Out]: The number of generated LTRs.

\item[Val]: The number of LTRs validated by the lexicographers
and thus added to the transfer lexicon.

\item[InOut]: The number of input word equivalences (\textbf{In}) for
which some output (\textbf{Out}) was provided. This value is usually
lower than the value of \textbf{Out} because more than one LTR can be
created for the same word equivalence. Therefore, more than one
element of \textbf{Out} can correspond to one element of
\textbf{InOut}.

\item[InVal]: The number of input word equivalences (\textbf{In}) for
which some LTR was validated (\textbf{Val}) by the lexicographers.

\item[\%]: The success rate, obtained by dividing the value of
\textbf{InVal} by the value of \textbf{In}. By using the value of
\textbf{InVal} instead of the value \textbf{Val} we factor out the
extra valid LTRs that can be created for a given input word
equivalence, in addition to the first one.

\end{description}

The listed files are sorted according to the chronological order in
which they were processed, and divided according to the methodology in
use. One of the most common reasons for failure is the presence of
unknown words on the English side. When the input contains unknown
words, we block generation. In contrast unknown words are accepted on
the Spanish side, matching any possible lexical category. This
treatment of Spanish unknown words is one of the reasons that explains
the higher value in \textbf{Out} than in \textbf{InOut}. Genuine
syntactic ambiguity is the other main reason for such difference in
values.

In terms of speed, we ran a test by evaluating the development time of
the file named `Phrasal verbs - batch 1' in Table
\ref{performance}. We had a lexicographer timing three activities:
coding translation pairs, revising automatically generated LTRs,
manually coding LTRs for the translation pairs that failed in
automatic generation. The results are shown in Table \ref{speed}.

\begin{table*}[tb]
\begin{center}
\begin{tabular}{llcc}
Activity			& N. of items		& Items per hour	& Time for 100 items	\\ \hline
Coding translation pairs	& 2340			& 31.25			& ~3h 12m		\\
Revising LTRs			& 1416 (validated)	& 50.57			& ~1h 59m		\\
Manually coding LTRs		& ~926			& ~6.25			& 16h 00m		\\
\end{tabular}
\end{center}
\caption{Speed test results.}\label{speed}
\end{table*}

\section{Conclusion}

The described methodology makes LTR coding considerably
faster. According to our test, revising automatically generated LTR
files is about 8 times as fast as manually coding LTRs. If the manual
coding of word equivalences is also counted in the automatic
generation process, the process is still about 3 times as fast as the
manual coding of complete LTRs.

Besides speed, this methodology guarantees the syntactic correctness
of the output (provided that the templates are syntactically correct,
of course). The validation procedure only requires removing unwanted
LTRs, with no further editing intervention. Also, more control over a
transfer rule database is provided, as each LTR can be associated with
a template. In this way testing, debugging, and maintenance in general
are easier and more effective, as these processes can be performed on
templates rather than on LTRs.

Although the described methodology can hardly aspire to completeness,
compared to manual coding, it can be integrated with a manual coding
phase, for the translation pairs that fail to generate
automatically. The gain in terms of labour effort is still
proportional to the success rate of the generation procedure, when
compared to an entirely manual coding. Moreover, there is one sense in
which automatic generation is more complete than manual coding, namely
in the generation of multiple entries for a translation
pair. Lexicographers tend to code one LTR per translation pair,
whereas the range of candidates proposed by automatic generation can
make them aware of valid alternatives analyses unnoticed by them.

Lexicographic work is easier, as little technical \linebreak knowledge
is required of lexicographers. Full knowledge of the formalism in use
for LTRs is only required in the initial phase of LTR corpus
constructions. Once a template generation procedure is in place,
lexicographers only need a passive knowledge of the formalism,
i.e. they must be able to read and understand LTRs, but not to write
LTRs.

Input word equivalences are expressed in plain natural language. This
makes them amenable to acquisition from corpora or MRDs. If manual
coding of word equivalences is necessary, only bilingual speaking
competence is required of lexicographers. No further linguistic
background or familiarity with any formalism is necessary.

The bootstrap approach makes this methodology more suitable for the
scaling up of large scale MT systems, than for rapid development of
prototypes. The knowledge acquired in prototype development is used in
developing a full-fledged system, which is often the most critical
phase for MT systems. The adoption of this methodology is also
profitable in porting an MT system to a different language pair or in
developing a multilingual MT system: the methodology is applicable to
any language pair, and the linguistic knowledge can also be re-used to
some extent, depending on the similarity between the languages at
hand.

\bibliographystyle{plain}

\bibliography{99mtsummit}

\end{document}